# Combining low-dose CT-based radiomics and metabolomics for early lung cancer screening support

Joanna Zyla[1,†] [ORCID: 0000-0002-2895-7969], Michal Marczyk[1,2,†,*] [ORCID: 0000-0003-2508-5736], Wojciech Prazuch[1,†], Marek Socha[1], Aleksandra Suwalska[1], Agata Durawa[3], Malgorzata Jelitto-Gorska[3], Katarzyna Dziadziuszko[3], Edyta Szurowska[3], Witold Rzyman[3], Piotr Widlak[3], Joanna Polanska[1] [ORCID: 0000-0001-8004-9864]

[1] Department of Data Science and Engineering, Silesian University of Technology, Gliwice, Poland

[2] Yale Cancer Center, Yale School of Medicine, New Haven, CT, USA

[3] Medical University of Gdansk, Gdansk, Poland

[†] Equal contribution

*Corresponding author: Michal Marczyk, michal.marczyk@polsl.pl, Department of Data Science and Engineering, Silesian University of Technology, Gliwice, Poland

**Abstract**: Due to its predominantly asymptomatic or mildly symptomatic progression, lung cancer is often diagnosed in advanced stages, resulting in poorer survival rates for patients. As with other cancers, early detection significantly improves the chances of successful treatment. Early diagnosis can be facilitated through screening programs designed to detect lung tissue tumors when they are still small, typically around 3mm in size. However, the analysis of extensive screening program data is hampered by limited access to medical experts. In this study, we developed a procedure for identifying potential malignant neoplastic lesions within lung parenchyma. The system leverages machine learning (ML) techniques applied to two types of measurements: low-dose Computed Tomography-based radiomics and metabolomics. Using data from two Polish screening programs, two ML algorithms were tested, along with various integration methods, to create a final model that combines both modalities to support lung cancer screening.

## INTRODUCTION

Lung cancer ranks as the top cause of cancer-related mortalities for both men and women. Among men, it accounts for 21% of all cancer deaths, followed by prostate (11%) and colon & rectum cancers (9%). In women, it represents 21% of all cancer deaths, followed by breast (15%) and colon & rectum cancers(8%) [1]. Approximately 81% of lung cancer deaths result from direct cigarette smoking, with indirect smoking being a secondary contributing factor. The diagnosis of lung cancer typically initiates with lung parenchyma screening, primarily conducted through chest radiography or Computed Tomography (CT). Radiologists play a crucial role in identifying nodules or tumors within the lung parenchyma. Subsequently, they assess the radiological characteristics of these detected objects to determine malignancy. If any nodule displays malignant features, the process of lung cancer diagnosis begins.

The low survival rate for lung cancer is mainly related to late diagnosis. An asymptomatic course of disease progression causes a low percentage diagnosis of lung cancer at an early stage of the disease. In the case of symptoms, long-time smoking patients have usually already developed chronic symptoms like the early signs of lung cancer and often disregard the symptoms if they are not severe. Patients' beliefs about health changes that may indicate lung cancer appeared to have played a part in the delay in diagnosis. Unfortunately, late admission and oncological therapy often fail to cure the patient. Still, they could do so in the earlier stages of the disease, because when the diagnosis occurs at a late stage, the survival rate is low [2].

Screening initiatives have been established to enhance the rate of early lung cancer detection in cases where the disease remains asymptomatic during its progression or exhibits mild, non-alarming symptoms to the patient. These trials target a high-risk demographic comprising middle-aged and elderly individuals with a history of long-term smoking. Ideally, such screenings should occur annually, which places a significant burden on current radiological resources. The evaluation of image-based screenings for detecting subtle abnormalities can be a complex and time-consuming process. To overcome these limitations, in this study, we developed a model for early lung cancer screening support by combining the results of low-dose CT and metabolomics modalities. We gathered data from two different screening cohorts and tested two machine-learning models and several integration algorithms to find the best solution.

## MATERIALS AND METHODS

### Study Subjects

Material included in this study was collected during the MOLTEST (between 2009 and 2011) and MOLTEST-BIS (between 2016 and 2017) Lung Cancer Screening Programs performed by the Medical University of Gdansk [3]. These programs enrolled several thousand participants and offered low-dose computed tomography (LDCT) examinations for current or former smokers with at least a 20-pack-year history, aged from 50 to 75 years. This report involves two groups of participants of the project: (i) patients with CT-detected lung nodules that were confirmed benign by histopathology further marked as benign, and (ii) patients who were ultimately diagnosed with lung cancer further marked as malignant. For those patients, two types of measurements (modalities) were collected: (i) metabolomic profiles from peripheral blood; (ii) radiomic characteristics from LDCT scans. The metabolomic data were collected for 246 patients from the MOLTEST-BIS cohort. The radiomic features were extracted from 1086 patients from both, MOLTEST and MOLTEST BIS datasets. Several regions of suspicious elements from the LDCT scan were collected for one of the patients. The basic characteristics of the analyzed data are presented in **Table 1**. Studies were approved by the appropriate Ethics Committees (Medical University of Gdansk, approval no. NKBBN/376/2014), and all participants provided informed consent indicating their voluntary participation in the project and provision of blood samples for future research.

**Table 1. Basic characteristics of the analyzed population.**

| | **Radiomics** | **Metabolomics** |
|---|---|---|
| **Benign** | | |
| n= | 994 (75%) | 123 (50%) |
| Sex: male/female | 445/549 (45%/55%) | 66/57 (54%/46%) |
| **Malignant** | | |
| n= | 331 (25%) | 123 (50%) |
| Sex: male/female | 136/195 (41%/59%) | 67/56 (54%/46%) |

**Metabolomics data**

The detailed procedure of data collection, preparation, and preprocessing is described elsewhere [4]**.** Briefly, for collected peripheral blood from 246 MOLTES-BIS patients, the measures were obtained from combined direct flow injection and liquid chromatography high-resolution mass spectrometry assay using the Absolute IDQ p400 HR kit (test plates in the 96-well format; Biocrates Life Sciences AG, Innsbruck, Austria) according to the manufacturer's protocol. For obtained measurements, the batch correction was performed due to several experiment runs. Finally, 259 metabolites were used in further analysis as their measures include less than 50% of signals below the level of detection/quantitation per patient. Moreover, for the following groups of metabolites, an average signal was obtained: acylcarnitines, amino acids, biogenic amines, glycerophospholipid, sphingolipids, cholesterol esters, glycerides, TG, DG, LPC, PC, and lipids. In total 271 features for metabolomics data were analyzed.

**Radiomic data**

Radiomics data for this project comes from measurements done on 1086 patients (925 from MOLTEST and 161 from MOLTEST-BIS). For each patient, the LDCT was performed and for abnormalities found in the lung tissue, an annotation was prepared by an expert radiologist. Regions with abnormalities were categorized into the following groups: suspicious nodules, inflammation, benign nodules, lymph nodes, fibrosis, and calcification. Both, suspicious nodules, and inflammation were considered malignant cases, the remaining categories represent the benign group. Also, one patient can have changes in lung tissue from several of the abovementioned groups. Next, the mask of objects and its segmentation were extracted using multi-step pre-processing and segmentation algorithm [5]**.** For extracted regions of the lung, the radiomics features were calculated with the usage of the PyRadiomics package [6]. In total, 107 radiomics features were analyzed for 5180 fragments of annotated images from analyzed patients. Finally, radiomics features were internally standardized (scaled and shifted) using non-parametric statistics like median and interquartile range calculated on benign samples within a cohort.

**Univariate analysis**

Each analyzed feature from both modalities was tested due to the normality of distribution using the Shapiro-Wilk test. As they are highly skewed to estimate the significance of differences in analysis groups, the Mann-Whitney test was used. Moreover, each feature's biserial correlation ($r_g$) was calculated and treated as an effect size measure. Finally, the Benjamini−Hochberg procedure for the FDR correction was applied when necessary [7]. All statistical hypotheses were tested at the 5% significance level.

**Machine learning sets**

The classification models were constructed using two different machine learning (ML) approaches: (i) logistic regression (LR), and (ii) random forest (RF). Both ML methods were performed on the same training and test sets. The test set was extracted from the MOLTEST-BIS cohort by taking 20 benign and 20 malignant cases for which both radiomic and metabolomic data were available. Moreover, for radiomic data, the test set was expanded for additional cases from the MOLTEST cohort. The remaining samples were gathered into a training set. A summary of the training and test set for both omics is presented in **Table 2**.

**Table 2. Number of samples in training and test sets for ML model building.**

| | **Radiomics** | **Metabolomics** | **Common** |
|---|---|---|---|
| **Train** | | | |
| **Benign** | 4569 | 103 | 68 |
| **Malignant** | 440 | 103 | 53 |
| **N** | 5009 | 206 | 121 |
| **Test** | | | |
| **Benign** | 122 | 20 | 20 |
| **Malignant** | 49 | 20 | 20 |
| **N** | 171 | 40 | 40 |

**Logistic regression models**

For the initially selected train set the Multiple Random Cross-Validation (MRCV) was performed as follows: (i) data split into training and validation (70%/30%) subsets; (ii) forward feature selection for logistic regression on training subset with stop criterion $\Delta BIC \leq 2$; (iii) evaluation of classification parameters on the train and validation subsets; (iv) estimation of classification threshold by maximizing Balanced Accuracy (BAcc). The MRCV procedure was repeated 100 times. Next, feature ranking was generated as follows: (i) features included in each model were sorted by their order of addition in the forward procedure; (ii) proportional order was multiplied by BAcc of the validation set at a particular fold; (iii) the elbow technique was used to extract the most relevant features for the final LR model. Finally, the model was built on the entire train set using selected features and evaluated on the test set.

**Random forest models**

Random forest classifier was implemented using *caret* R package [8] with sample weighting to decrease the effect of class imbalance. Two RF model parameters were tested: (i) Mtry – number of features sampled at each tree split (from 5 to 30); (ii) Ntree – number of trees in a forest (100, 500, 1000, 2000). MRCV procedure was applied as follows: (i) data split into training and validation (80%/20%) subsets; (ii) RF model fit on training data; (iii) estimation of variable importance (for each tree, the prediction accuracy on the out-of-bag portion of the data is recorded; then the same is done after permuting each predictor variable; the difference between the two accuracies is then averaged over all trees, and normalized by the standard error); (iv) estimation of classification threshold by maximizing Balanced Accuracy. The MRCV procedure was repeated 100 times. Next, feature ranking was generated as follows: (i) for each feature, the average variable importance score was calculated; (ii) the elbow technique was used to extract the most relevant features for the final RF model. Finally, the model was built on the entire train set using selected features and evaluated on the test set.

**Machine learning results integration**

Several approaches were applied to integrate results from analyzed omics platforms. The first one was based on statistical integration proposed by Stouffer [1]. It was used for both, classification probabilities of test sets as well as classification thresholds. Additionally, several common methods were tested including: (i) mean value; (ii) maximum value; and (iii) product of two probabilities. Similarly, both classification probabilities and classification thresholds were integrated using the same method.

**RESULTS**

To construct classification models, we used two different cohorts, MOLTEST and MOLTEST BIS, and two types of measurements, radiomic features from LDCT and metabolomic data. We cleaned the datasets by removing missing values and data normalization. For both modalities, there is no difference between cohorts, but only after data normalization (**Figure 1**). In radiomics, there is a better separation between benign and malignant cases than in metabolomics, but these two classes are not separated. Also, we observe three clusters of data points in radiomic data, and smaller clusters are mostly formed with benign cases (**Figure 1B**). By further investigation, we found that in one of the clusters, there are mainly calcified nodules, which resulted from historical infections or physical damages. The smallest cluster consists of all nodule types categorized into both classes, so it might represent some unknown technical differences.

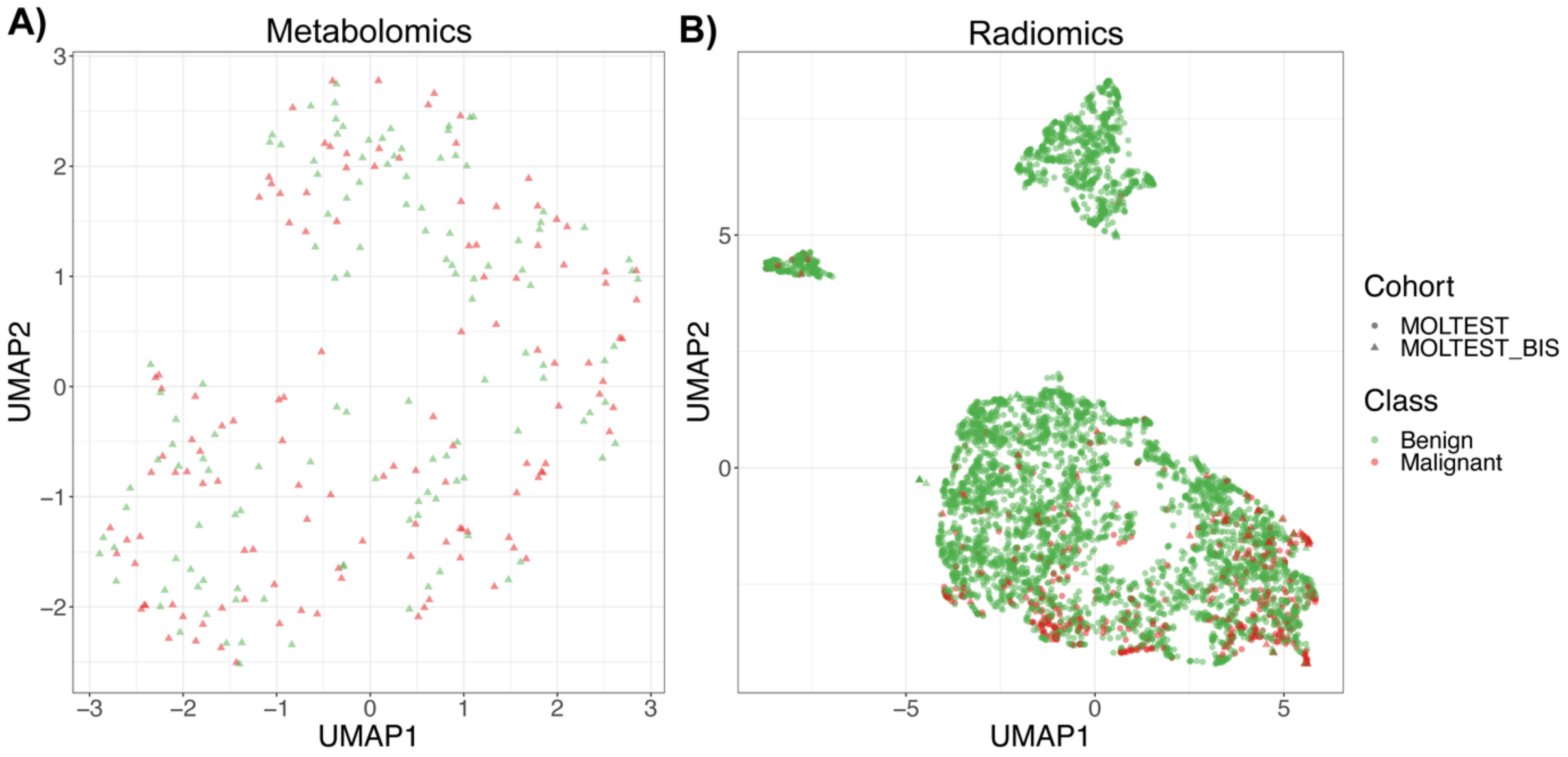

**Figure 1. UMAP visualization of all patients in the metabolomics (A) and radiomics (B) studies.**

**Univariate analysis of metabolomic and radiomic studies**

Many features measured in both modalities were highly skewed, showing non-normal distribution (**Table S1**), so we used non-parametric methods for univariate analysis to compare benign and malignant cases within each modality. We run the analysis using all available samples in each modality. After applying multiple testing corrections, we found no significant features in the metabolomics study and 94 significant features in the radiomics study, from which 44 were downregulated and 50 upregulated. The effect size measure for top differentially regulated features was also higher in radiomics than in metabolomics (0.6 vs 0.26; **Figure 2**). Many features showed similar patterns of feature level difference, due to the high correlation between features within studies (**Figure S1**).

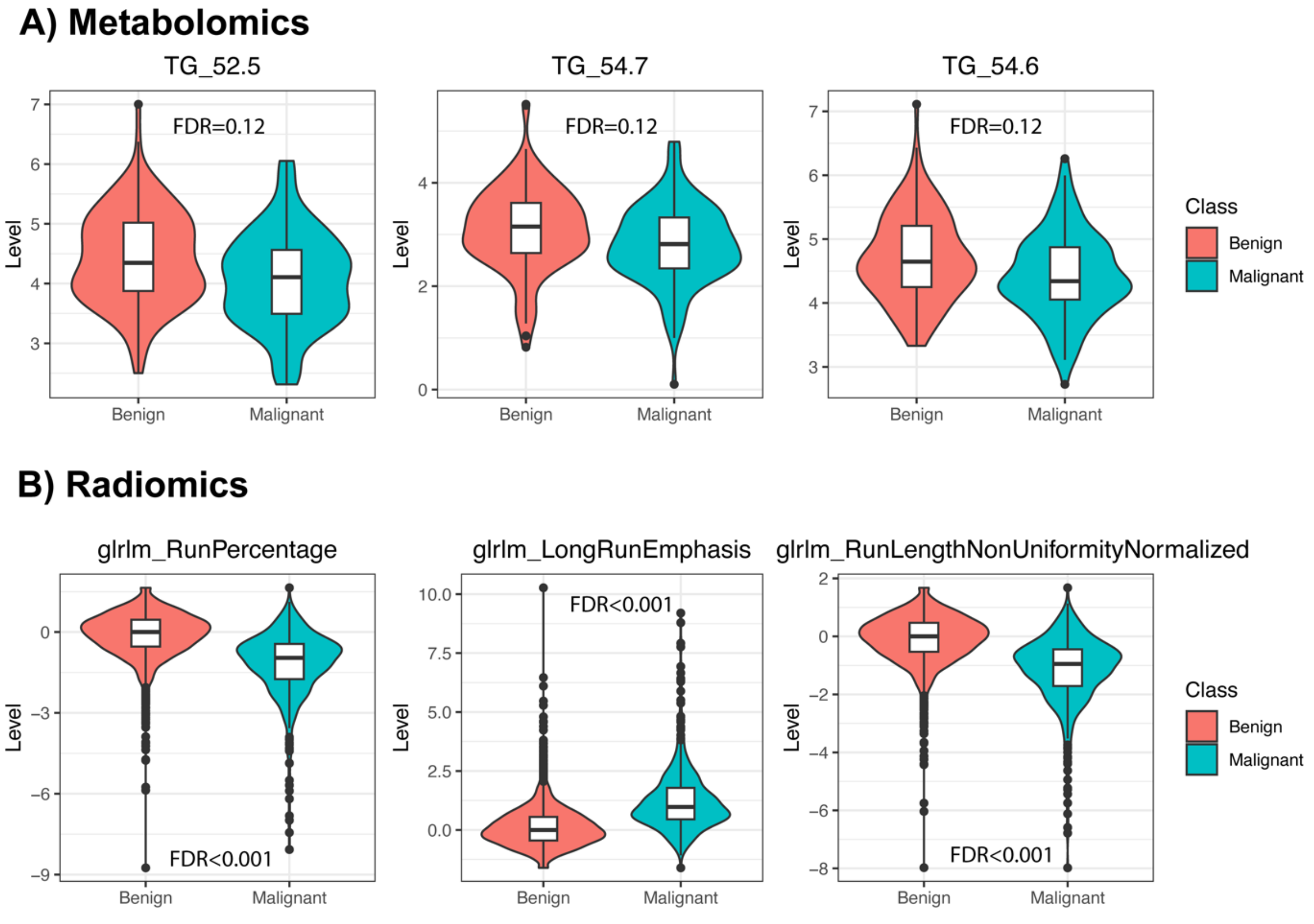

**Figure 2. Top 3 most differentially changed features in metabolomics (A) and radiomics (B) modalities.**

### Development of machine learning models

Before modeling, we removed highly correlated features in each modality. MRCV procedure used for parameter tuning and selection of important genes (**Figures S2, S3**) resulted in a different number of features (Metabolomics: 11 for LR and 14 for RF; Radiomics: 11 for LR and 8 for RF). Only 3 features were common between ML models in metabolomics (*PC_41.5*, *PC_O_42.6*, and *PC_42.7*) and 3 in radiomics (*glcm_InverseVariance*, *shape_Flatness*, and *glcm_Id*). More than half of the patients in the test set were properly classified by all models after integration (**Figure 3**). In most cases, integration decreases the number of false positive and negative findings. Similar findings were observed in the training set (**Figures S4, S5**).

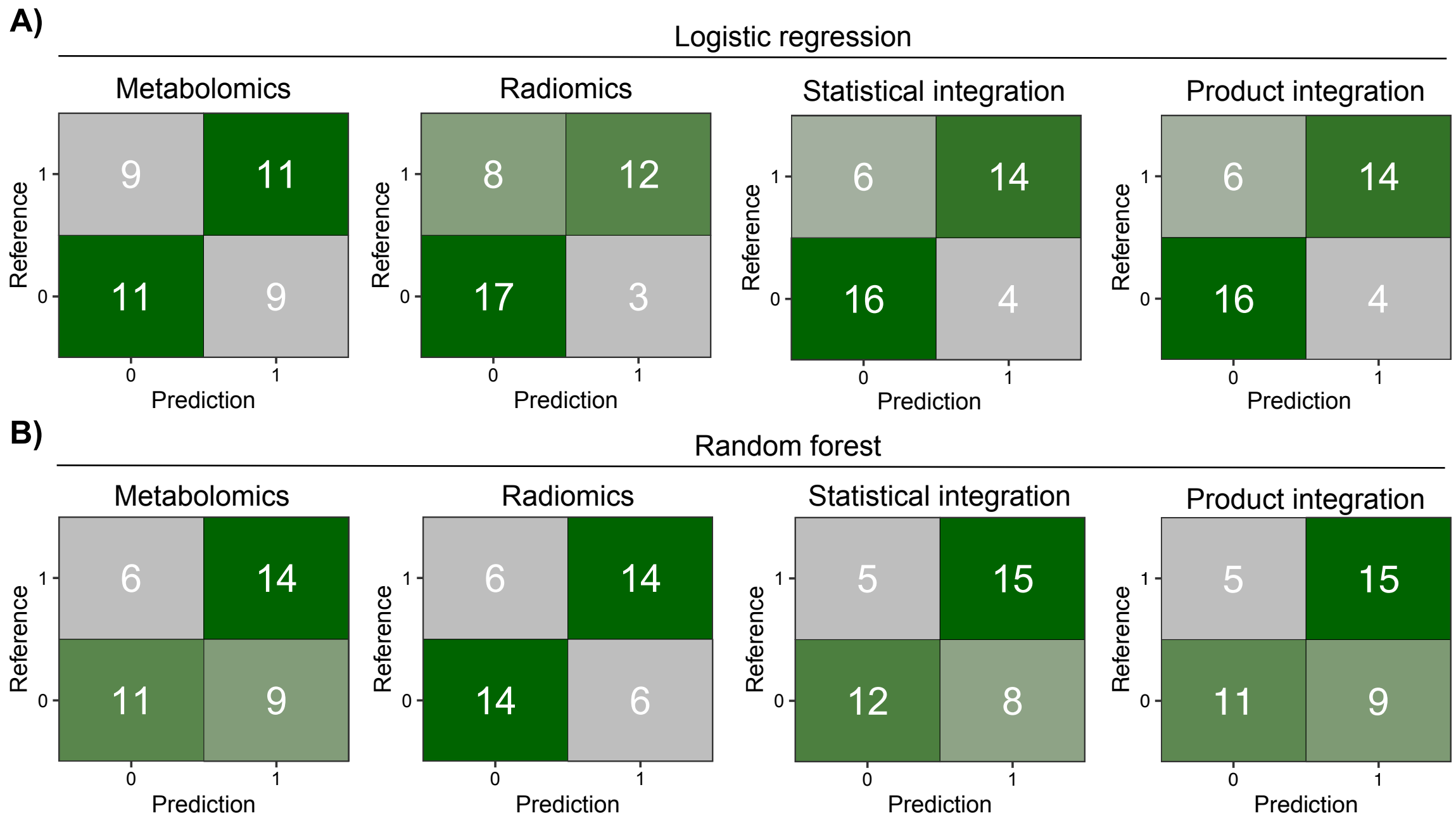

**Figure 3. Heatmaps of confusion matrices for prediction models on the MOLTEST-BIS test set for each modality and after integration.** (A) Results of logistic regression models. (B) Results of random forest models.

Looking at the model performance indices on the test set after applying the estimated classification thresholds we observed diverse results. RF was better than LR for metabolomics data, however, it was worse for radiomics and after integration (**Table 3**). The overall performance was moderate with F1 ranging from 0.55 to 0.8 and AUC from 55.5% to 84.8%. ROC curve analysis showed that there is a potential to tune threshold values to meet other specific goals of a model (**Figure 4**). Results on the training set showed that models were not significantly overtrained (**Table S2, Figure S6**). Integrating classification results from two modalities increases model performance in comparison to basic models (e.g., F1 and BAcc) regardless of the integration method used (**Table 3**).

**Table 3. Results for ML models on MOLTEST-BIS test set. The bold** value shows the superiority of a particular metric and set of data between MLS approaches; LR - logistic regression, and RF- random forest.

| Metric | Radiomics | | Metabolomics | | Statistical integration | | Product integration | |
|---|---|---|---|---|---|---|---|---|
| | LR | RF | LR | RF | LR | RF | LR | RF |
| **Sensitivity** | 0.60 | **0.70** | 0.55 | **0.70** | 0.70 | **0.75** | **0.78** | 0.63 |
| **Specificity** | **0.85** | 0.70 | **0.55** | **0.55** | **0.80** | 0.60 | **0.73** | 0.69 |
| **PPV** | **0.80** | 0.70 | 0.55 | **0.61** | **0.78** | 0.65 | 0.7 | **0.75** |
| **NPV** | 0.68 | **0.70** | 0.55 | **0.65** | **0.73** | 0.71 | **0.8** | 0.55 |
| **F1** | 0.69 | **0.70** | 0.55 | **0.65** | **0.74** | 0.70 | **0.74** | 0.68 |
| **Balanced Accuracy** | **0.73** | 0.70 | 0.55 | **0.63** | **0.75** | 0.68 | **0.75** | 0.66 |
| **AUC** | **83.0%** | 70.3% | 55.5% | **60.3%** | **83.0%** | 73.0% | **84.8%** | 71.5% |

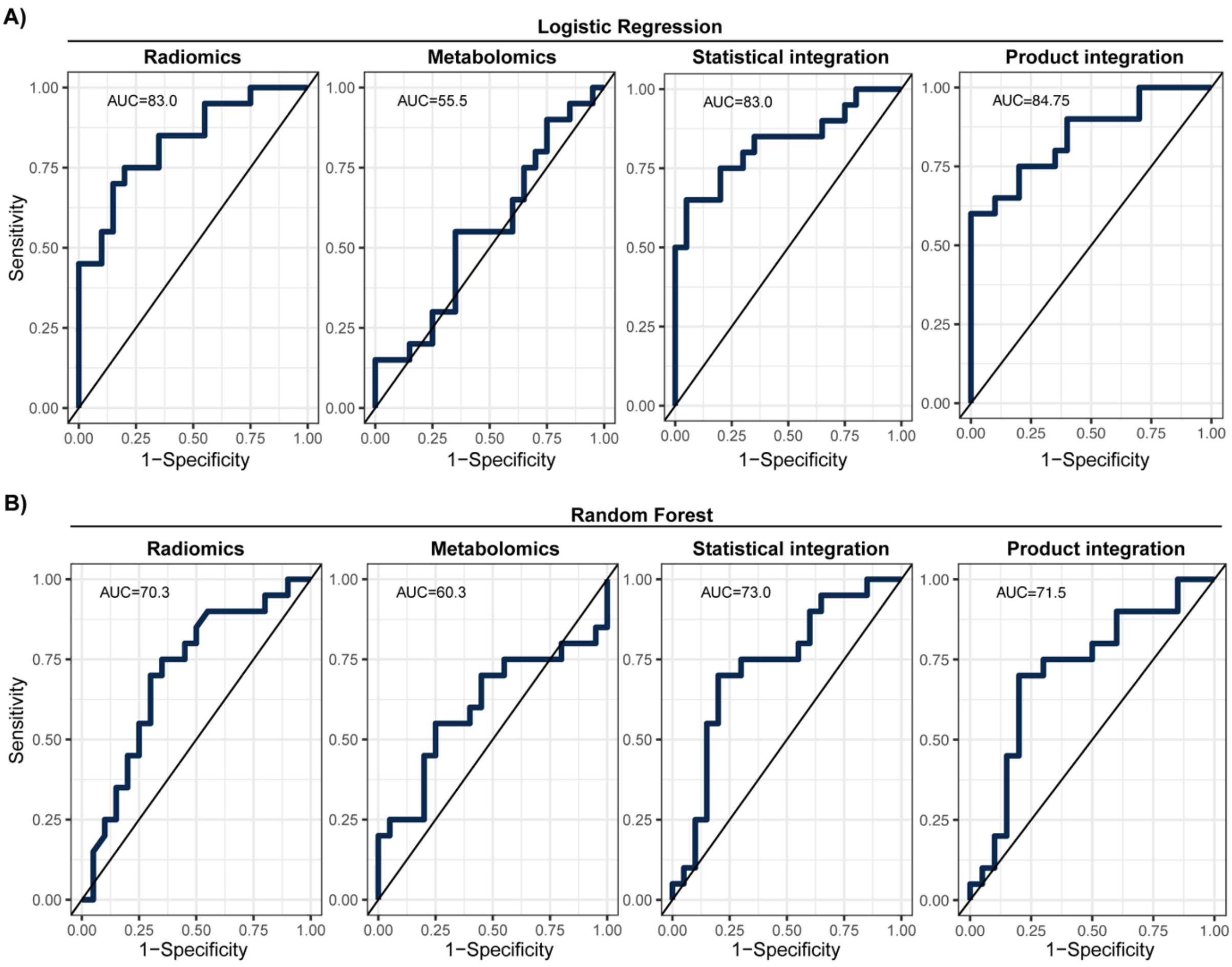

**Figure 4. ROC with given AUC for prediction models on the MOLTEST-BIS test set for each modality and after integration.** (A) Results of logistic regression models. (B) Results of random forest models.

**Funding**

This research was funded by the National Science Centre, Poland, grant 2017/27/B/NZ7/01833. Clinical material was collected and initially characterized in the frame of project MOLTEST-BIS (DZP/PBS3/247184/2014). The study was partially supported by the Silesian University of Technology grant for Support and Development of Research Potential. All research performed by MM, WP, JZ, and JP is within the scope of the Technical Informatic and Telecommunication discipline recognized by the Polish Ministry of Science.

## Supplementary materials

## *Combining low-dose CT-based radiomics and metabolomics for early lung cancer screening support*

**Table S2. Results for ML models on train and test set (without removal of MOLTEST-2013 from test set). The bold** value shows the superiority of a particular metric and set of data between models; LR - logistic regression, RF- random forest.

| | **Radiomics** | | | | **Metabolomics** | | | |
|---|---|---|---|---|---|---|---|---|
| **Metric** | **Train** | | **Test** | | **Train** | | **Test** | |
| | **LR** | **RF** | **LR** | **RF** | **LR** | **RF** | **LR** | **RF** |
| **Sensitivity** | **0.78** | 0.67 | **0.69** | **0.69** | **0.76** | 0.61 | 0.55 | **0.70** |
| **Specificity** | 0.78 | **0.82** | **0.74** | 0.68 | **0.74** | 0.66 | 0.55 | **0.55** |
| **PPV** | 0.25 | **0.27** | **0.51** | 0.47 | **0.74** | 0.64 | 0.55 | **0.61** |
| **NPV** | **0.97** | 0.96 | **0.86** | 0.85 | **0.75** | 0.62 | 0.55 | **0.65** |
| **F1** | **0.38** | **0.38** | **0.59** | 0.56 | **0.75** | 0.62 | 0.55 | **0.65** |
| **Balanced Accuracy** | **0.78** | 0.74 | **0.72** | 0.69 | **0.75** | 0.63 | 0.55 | **0.63** |
| **AUC** | **85.2%** | 83.2% | **76.7%** | 72.6% | **78.9%** | 66.4 | 55.5% | **60.3%** |

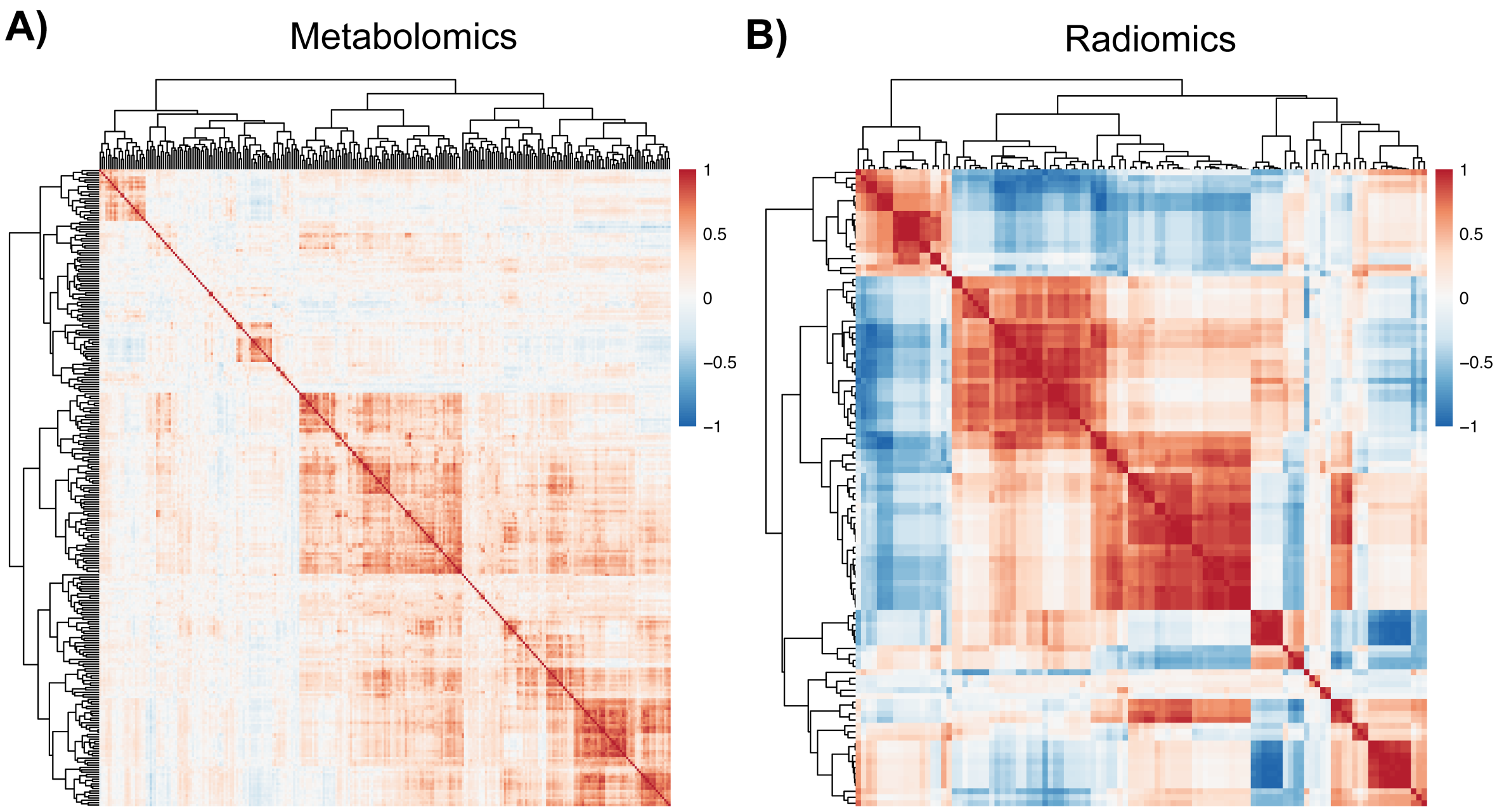

**Figure S1. Spearman correlation between features in metabolomics (A) and radiomics (B) modalities.**

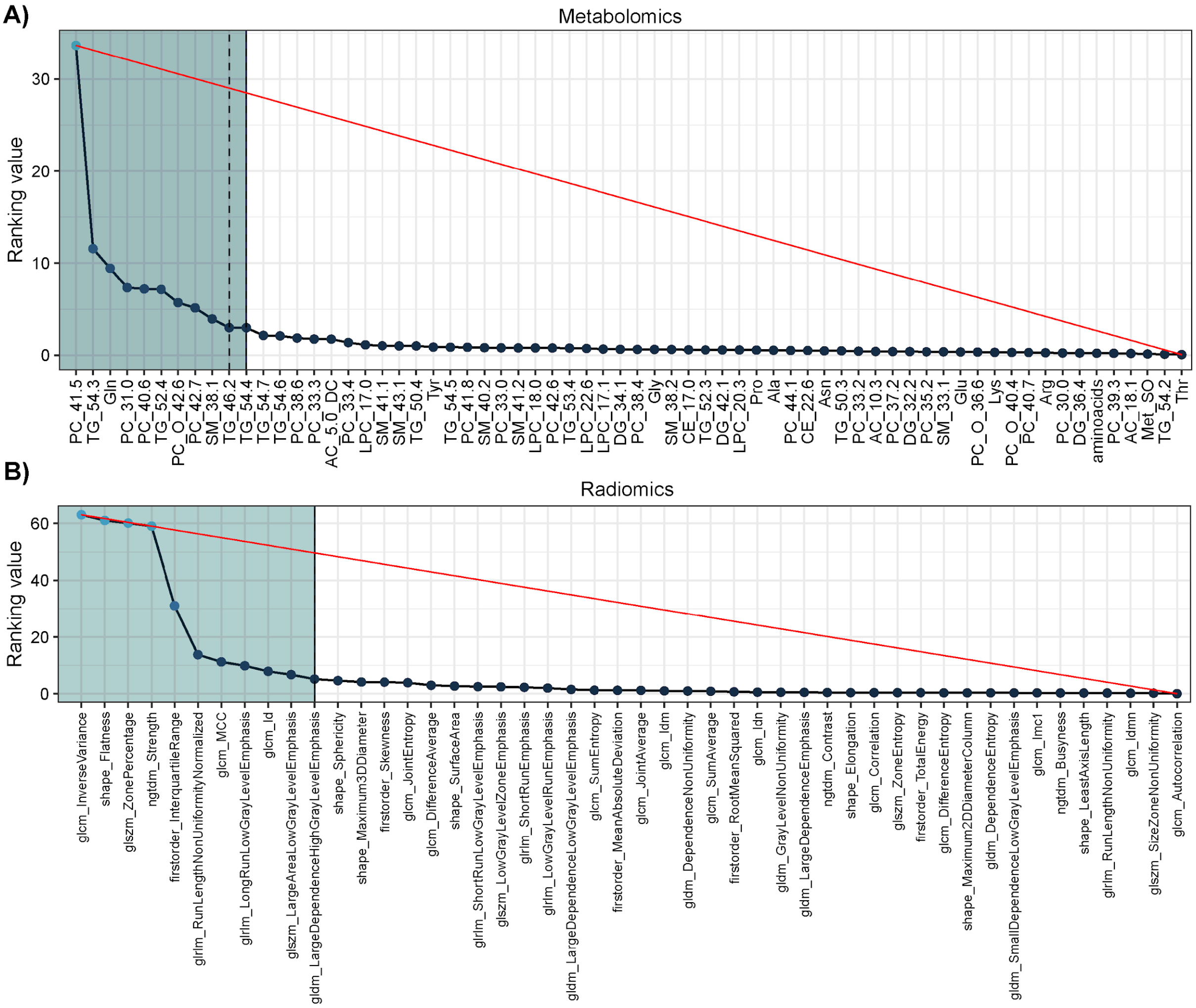

**Figure S2. Elbow plot of feature ranking for logistic regression approach.** Panels A and B show results for metabolomics and radiomics respectively. Pale blue are mark features taken to the final classifier.

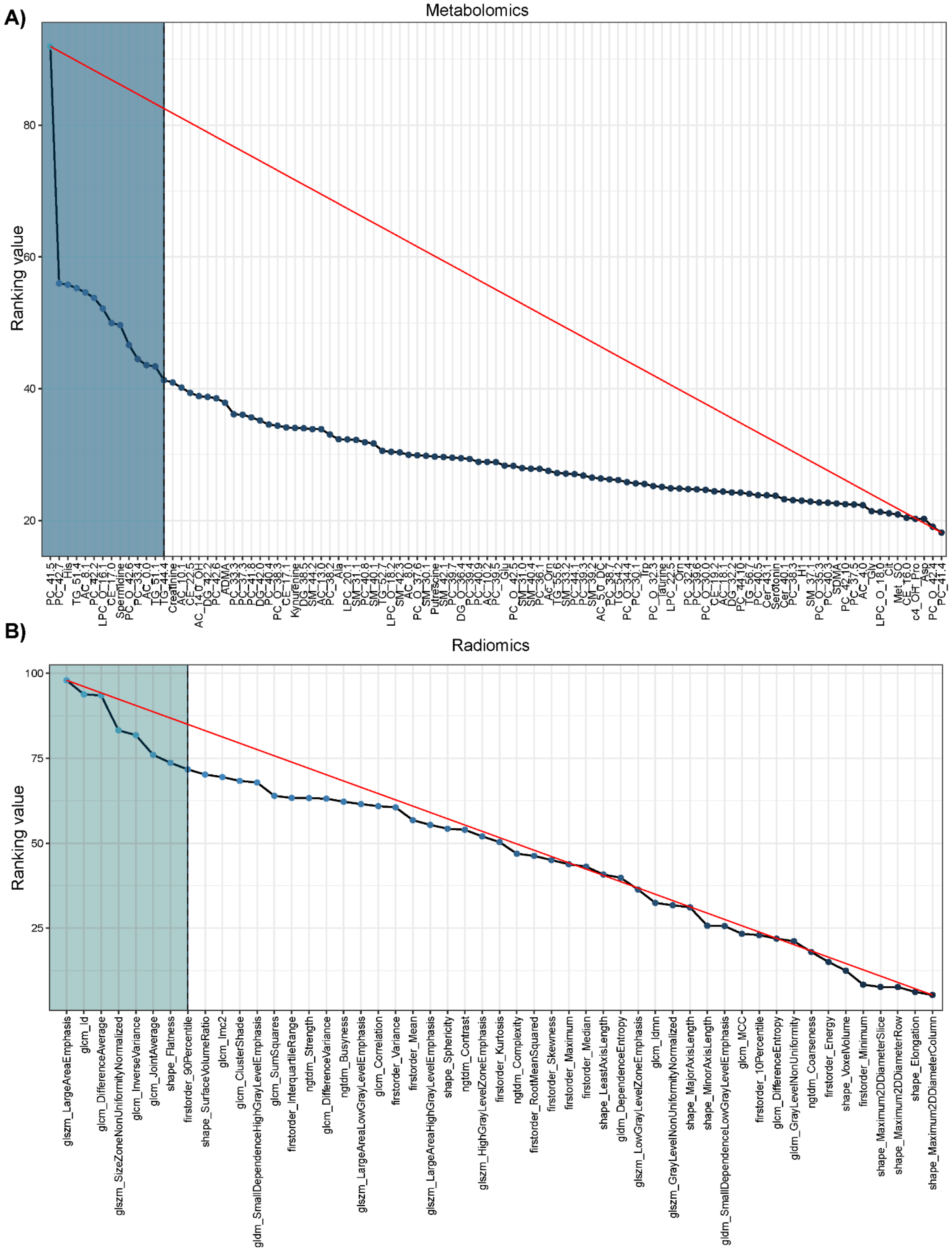

**Figure S3. Elbow plot of feature ranking for random forest approach.** Panels A and B show results for metabolomics and radiomics respectively. Pale blue are mark features taken to the final classifier.

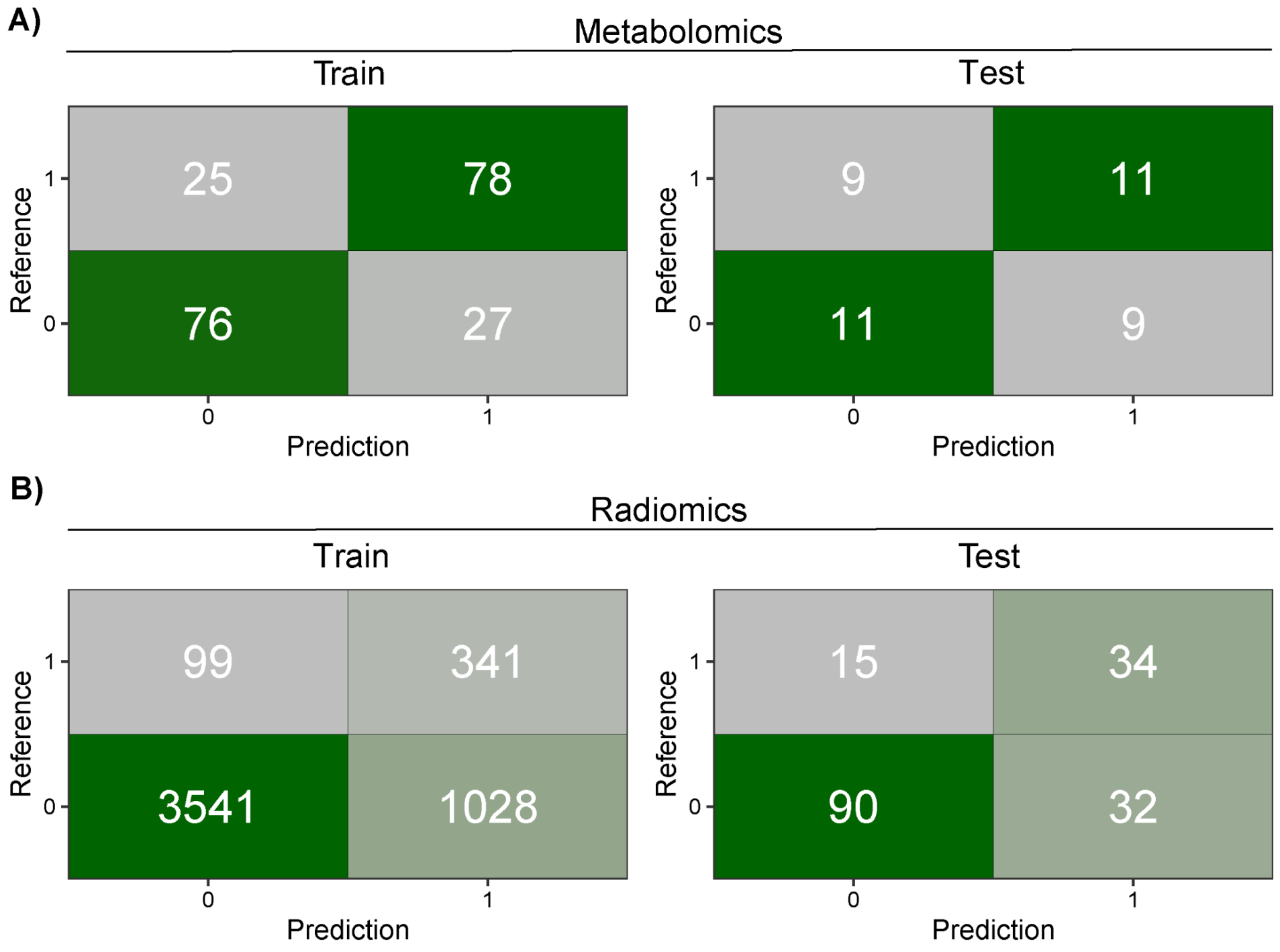

**Figure S4. Heatmaps of confusion matrices for LR models on the train and full test sets for metabolomics (A) and radiomics (B).**

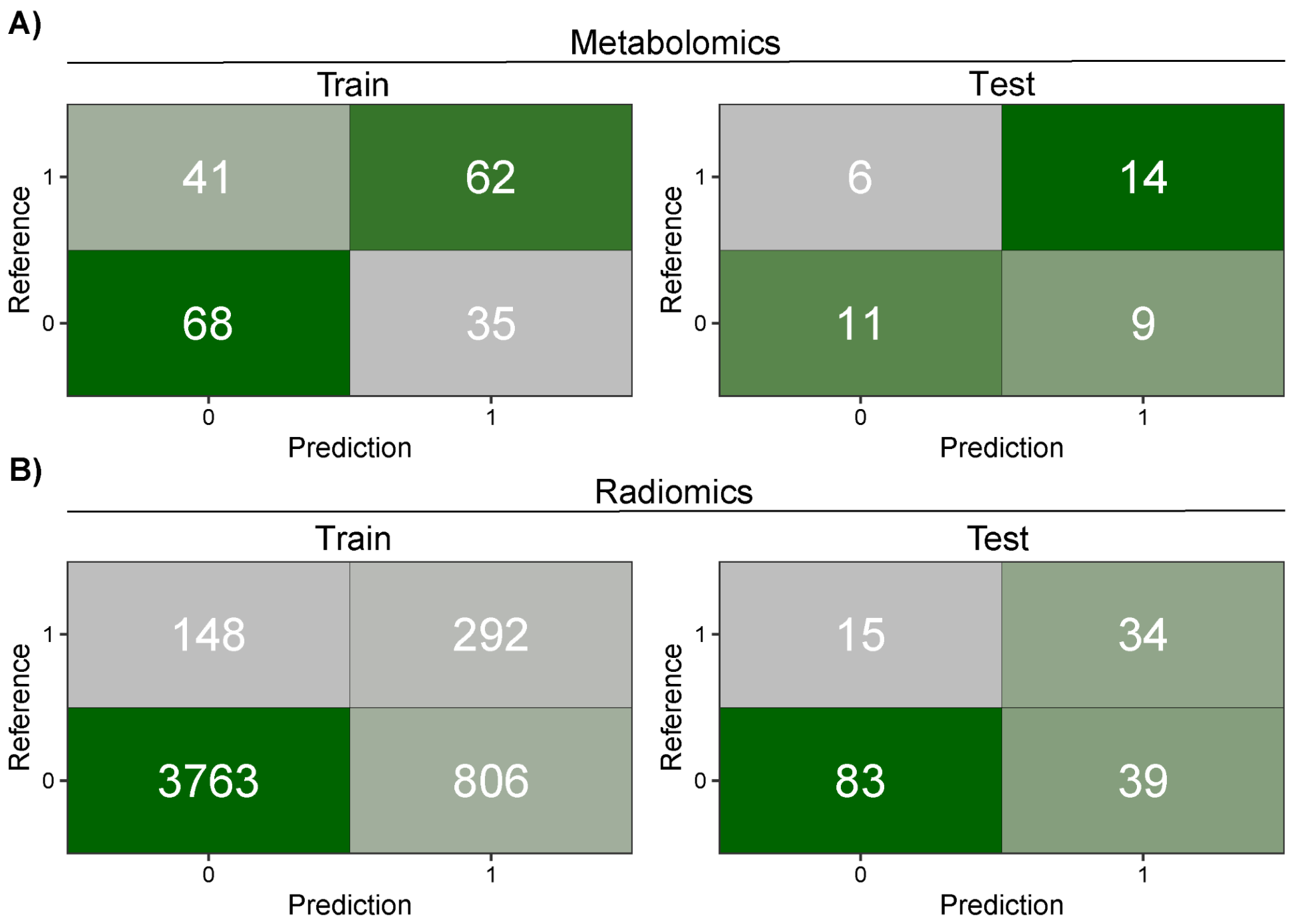

**Figure S5. Figure S4. Heatmaps of confusion matrices for RF models on the train and full test sets for metabolomics (A) and radiomics (B).**

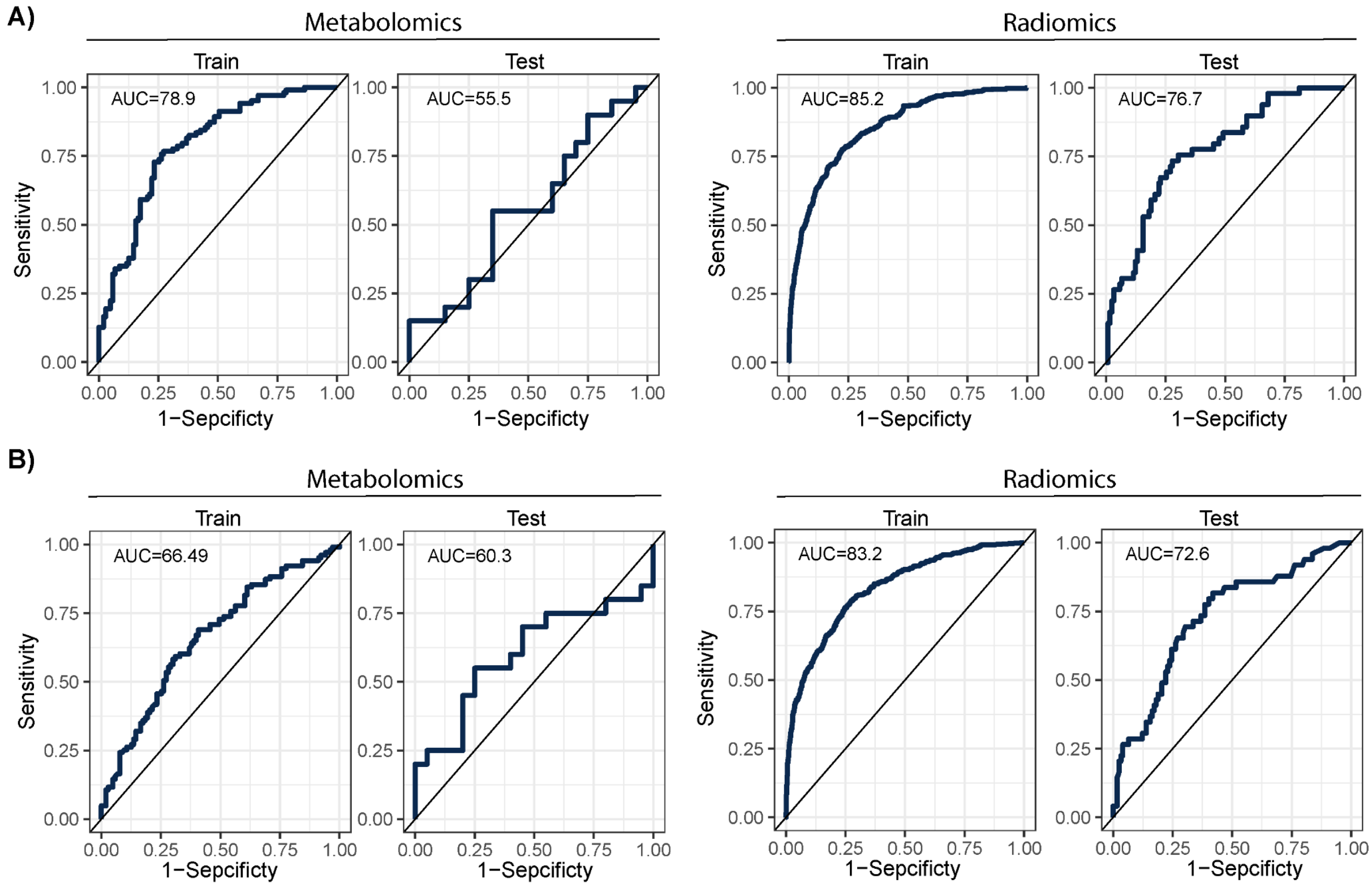

**Figure S6. ROC with given AUC for training set, test set and each modality.** Panel A shows results for logistic regression; Panel B shows results for random forest.

| Feature | Normality benign | Normality malignant | Rg effect size | P-value | FDR |
|---|---|---|---|---|---|
| TG_52.5 | 3.67E-01 | 3.34E-01 | -0.257 | 4.88E-04 | 1.20E-01 |
| TG_54.7 | 7.80E-02 | 1.66E-01 | -0.236 | 1.40E-03 | 1.20E-01 |
| TG_54.6 | 1.48E-01 | 7.46E-01 | -0.231 | 1.74E-03 | 1.20E-01 |
| TG_52.4 | 9.90E-01 | 9.03E-01 | -0.231 | 1.77E-03 | 1.20E-01 |
| SM_38.1 | 4.98E-01 | 2.74E-02 | -0.222 | 2.64E-03 | 1.24E-01 |
| PC_40.6 | 2.71E-02 | 8.18E-01 | -0.215 | 3.53E-03 | 1.24E-01 |
| glycerides | 7.29E-02 | 1.83E-01 | -0.212 | 4.03E-03 | 1.24E-01 |
| TG | 9.93E-02 | 3.29E-01 | -0.212 | 4.12E-03 | 1.24E-01 |
| TG_52.3 | 6.77E-01 | 5.91E-01 | -0.207 | 5.12E-03 | 1.24E-01 |
| TG_50.3 | 5.24E-01 | 4.56E-02 | -0.206 | 5.18E-03 | 1.24E-01 |
| PC_38.6 | 6.84E-01 | 1.06E-01 | -0.204 | 5.63E-03 | 1.24E-01 |
| TG_46.2 | 4.34E-05 | 1.10E-07 | -0.201 | 6.42E-03 | 1.24E-01 |
| DG | 1.66E-02 | 1.87E-01 | -0.199 | 7.00E-03 | 1.24E-01 |
| PC_41.5 | 2.14E-05 | 8.62E-09 | 0.198 | 7.34E-03 | 1.24E-01 |
| TG_54.5 | 8.10E-02 | 7.04E-01 | -0.197 | 7.62E-03 | 1.24E-01 |
| TG_50.1 | 6.50E-01 | 8.66E-02 | -0.197 | 7.66E-03 | 1.24E-01 |
| TG_50.2 | 5.70E-01 | 5.21E-01 | -0.195 | 8.08E-03 | 1.24E-01 |
| PC_42.7 | 1.31E-01 | 2.52E-01 | -0.194 | 8.48E-03 | 1.24E-01 |
| TG_52.2 | 4.37E-01 | 7.01E-01 | -0.194 | 8.70E-03 | 1.24E-01 |
| TG_50.4 | 6.28E-01 | 1.42E-01 | -0.192 | 9.22E-03 | 1.25E-01 |
| TG_52.6 | 3.76E-02 | 2.57E-04 | -0.189 | 1.03E-02 | 1.33E-01 |
| SM_41.1 | 8.09E-01 | 8.32E-01 | -0.183 | 1.30E-02 | 1.43E-01 |
| TG_54.4 | 2.20E-01 | 8.20E-01 | -0.183 | 1.33E-02 | 1.43E-01 |
| TG_54.3 | 1.93E-01 | 3.10E-01 | -0.179 | 1.50E-02 | 1.43E-01 |
| DG_36.3 | 9.54E-01 | 7.38E-01 | -0.179 | 1.51E-02 | 1.43E-01 |
| CE_22.6 | 5.22E-01 | 2.99E-01 | -0.179 | 1.52E-02 | 1.43E-01 |
| lipids | 8.21E-01 | 2.48E-01 | -0.179 | 1.54E-02 | 1.43E-01 |
| TG_51.2 | 7.62E-01 | 4.88E-01 | -0.177 | 1.62E-02 | 1.43E-01 |
| PC_O_38.6 | 8.62E-01 | 2.03E-01 | -0.177 | 1.64E-02 | 1.43E-01 |
| TG_51.4 | 2.57E-02 | 3.99E-01 | -0.177 | 1.64E-02 | 1.43E-01 |
| DG_34.1 | 9.88E-02 | 7.15E-01 | -0.177 | 1.66E-02 | 1.43E-01 |
| PC_38.5 | 9.22E-02 | 1.49E-01 | -0.176 | 1.72E-02 | 1.43E-01 |
| DG_36.4 | 4.92E-01 | 3.74E-01 | -0.175 | 1.79E-02 | 1.43E-01 |
| PC_O_40.7 | 7.62E-01 | 1.21E-02 | -0.175 | 1.79E-02 | 1.43E-01 |
| SM_38.2 | 3.43E-01 | 3.16E-01 | -0.173 | 1.88E-02 | 1.46E-01 |
| TG_56.7 | 7.39E-01 | 3.60E-02 | -0.172 | 1.95E-02 | 1.47E-01 |
| TG_56.6 | 3.00E-02 | 2.06E-01 | -0.170 | 2.16E-02 | 1.58E-01 |
| TG_48.2 | 2.90E-01 | 3.74E-02 | -0.168 | 2.27E-02 | 1.62E-01 |
| DG_36.2 | 2.99E-01 | 1.30E-01 | -0.167 | 2.36E-02 | 1.64E-01 |
| SM_42.2 | 3.66E-01 | 8.49E-01 | -0.164 | 2.60E-02 | 1.74E-01 |
| SM_41.2 | 1.31E-01 | 6.62E-01 | -0.164 | 2.64E-02 | 1.74E-01 |
| LPC_18.0 | 6.55E-01 | 3.90E-01 | -0.161 | 2.91E-02 | 1.84E-01 |
| TG_48.1 | 4.07E-07 | 4.61E-05 | -0.161 | 2.92E-02 | 1.84E-01 |
| LPC_O_16.1 | 2.25E-02 | 6.97E-02 | -0.158 | 3.22E-02 | 1.98E-01 |
| CE_20.5 | 2.74E-01 | 3.65E-02 | -0.158 | 3.28E-02 | 1.98E-01 |
| PC_33.4 | 3.63E-01 | 1.20E-01 | -0.155 | 3.62E-02 | 2.11E-01 |
| DG_39.0 | 2.13E-01 | 4.30E-02 | -0.153 | 3.86E-02 | 2.11E-01 |
| Gln | 2.03E-06 | 7.35E-02 | 0.152 | 3.88E-02 | 2.11E-01 |
| SM_39.1 | 6.34E-01 | 5.13E-01 | -0.152 | 3.88E-02 | 2.11E-01 |
| TG_48.3 | 6.31E-02 | 9.12E-04 | -0.152 | 3.93E-02 | 2.11E-01 |
| TG_51.3 | 4.18E-01 | 1.94E-01 | -0.152 | 3.98E-02 | 2.11E-01 |
| SM_43.1 | 5.74E-02 | 1.55E-01 | -0.150 | 4.16E-02 | 2.11E-01 |
| DG_41.1 | 4.34E-02 | 5.14E-01 | -0.150 | 4.21E-02 | 2.11E-01 |
| Ala | 6.82E-01 | 9.51E-04 | -0.150 | 4.27E-02 | 2.11E-01 |
| DG_42.1 | 2.33E-01 | 3.64E-01 | -0.148 | 4.47E-02 | 2.11E-01 |
| DG_O_32.2 | 1.35E-05 | 2.94E-01 | -0.148 | 4.53E-02 | 2.11E-01 |
| cerophospholip | 6.52E-01 | 5.64E-01 | -0.147 | 4.61E-02 | 2.11E-01 |
| PC | 6.52E-01 | 5.64E-01 | -0.147 | 4.61E-02 | 2.11E-01 |
| TG_53.3 | 2.58E-01 | 6.27E-01 | -0.147 | 4.71E-02 | 2.11E-01 |
| sphingolipids | 4.95E-01 | 3.07E-01 | -0.147 | 4.71E-02 | 2.11E-01 |
| SM_44.1 | 2.19E-06 | 1.71E-02 | -0.146 | 4.75E-02 | 2.11E-01 |
| PC_36.5 | 2.71E-01 | 9.96E-01 | -0.145 | 4.89E-02 | 2.11E-01 |
| DG_38.5 | 4.88E-05 | 5.99E-05 | -0.145 | 4.99E-02 | 2.11E-01 |
| Cer_42.1 | 1.02E-01 | 1.72E-01 | -0.144 | 5.04E-02 | 2.11E-01 |
| PC_40.4 | 3.77E-01 | 4.39E-01 | -0.144 | 5.06E-02 | 2.11E-01 |
| PC_O_36.6 | 2.73E-01 | 7.06E-01 | -0.141 | 5.56E-02 | 2.28E-01 |
| PC_O_40.6 | 9.94E-01 | 4.48E-01 | -0.140 | 5.87E-02 | 2.37E-01 |
| PC_40.9 | 1.33E-02 | 3.26E-01 | -0.138 | 6.09E-02 | 2.43E-01 |
| LPC_O_18.1 | 3.53E-01 | 1.81E-01 | -0.138 | 6.19E-02 | 2.43E-01 |
| DG_O_34.1 | 4.64E-02 | 1.61E-02 | -0.137 | 6.36E-02 | 2.46E-01 |
| LPC_22.6 | 7.58E-01 | 3.09E-01 | -0.135 | 6.76E-02 | 2.55E-01 |
| PC_32.3 | 6.78E-01 | 1.74E-01 | -0.135 | 6.78E-02 | 2.55E-01 |
| PC_42.2 | 1.38E-10 | 6.62E-11 | -0.131 | 7.66E-02 | 2.85E-01 |
| SM_33.1 | 9.06E-01 | 4.22E-01 | -0.130 | 7.88E-02 | 2.88E-01 |
| Pro | 4.39E-01 | 2.43E-01 | -0.128 | 8.22E-02 | 2.94E-01 |
| Creatinine | 2.09E-01 | 4.51E-10 | -0.128 | 8.25E-02 | 2.94E-01 |
| AC_5.0_DC | 1.22E-03 | 1.55E-03 | 0.127 | 8.57E-02 | 2.97E-01 |
| PC_41.8 | 6.50E-04 | 2.23E-05 | -0.127 | 8.57E-02 | 2.97E-01 |
| PC_40.1 | 4.07E-07 | 2.51E-06 | -0.126 | 8.67E-02 | 2.97E-01 |
| PC_O_40.8 | 9.35E-01 | 9.20E-02 | -0.125 | 9.04E-02 | 3.01E-01 |
| SM_42.3 | 2.77E-10 | 2.24E-10 | -0.125 | 9.04E-02 | 3.01E-01 |
| PC_O_36.5 | 8.87E-01 | 3.68E-01 | -0.124 | 9.17E-02 | 3.01E-01 |
| PC_O_38.5 | 5.15E-01 | 6.49E-02 | -0.124 | 9.21E-02 | 3.01E-01 |
| SM_36.1 | 9.75E-01 | 4.44E-01 | -0.123 | 9.63E-02 | 3.11E-01 |
| SM_44.2 | 4.89E-02 | 1.17E-01 | -0.122 | 9.92E-02 | 3.16E-01 |
| LPC_20.3 | 2.59E-01 | 4.72E-01 | -0.120 | 1.04E-01 | 3.27E-01 |
| holesterol este | 8.36E-02 | 4.48E-01 | -0.119 | 1.07E-01 | 3.27E-01 |
| Cer_42.2 | 2.38E-01 | 3.75E-01 | -0.119 | 1.07E-01 | 3.27E-01 |
| DG_34.3 | 1.35E-03 | 2.48E-01 | -0.118 | 1.09E-01 | 3.27E-01 |
| CE_17.0 | 5.48E-03 | 3.24E-02 | -0.118 | 1.10E-01 | 3.27E-01 |
| TG_49.2 | 4.82E-03 | 6.61E-02 | -0.118 | 1.10E-01 | 3.27E-01 |
| Trp | 3.86E-05 | 9.99E-03 | -0.115 | 1.19E-01 | 3.52E-01 |
| PC_34.5 | 3.35E-04 | 4.10E-01 | -0.113 | 1.25E-01 | 3.66E-01 |
| CE_18.3 | 3.15E-01 | 4.74E-01 | -0.112 | 1.30E-01 | 3.71E-01 |
| PC_O_34.1 | 5.15E-01 | 1.79E-02 | -0.112 | 1.30E-01 | 3.71E-01 |
| CE_18.2 | 1.67E-01 | 9.71E-01 | -0.111 | 1.34E-01 | 3.76E-01 |
| Kynurenine | 7.36E-03 | 9.06E-07 | -0.110 | 1.36E-01 | 3.76E-01 |
| DG_32.1 | 9.48E-02 | 1.16E-01 | -0.110 | 1.36E-01 | 3.76E-01 |
| acylcarnitines | 3.45E-01 | 1.71E-03 | -0.108 | 1.42E-01 | 3.84E-01 |
| CE_22.5 | 1.62E-10 | 9.03E-02 | -0.108 | 1.43E-01 | 3.84E-01 |
| TG_52.7 | 2.59E-02 | 3.33E-02 | -0.107 | 1.46E-01 | 3.84E-01 |
| PC_O_36.3 | 4.07E-01 | 3.19E-01 | -0.107 | 1.46E-01 | 3.84E-01 |
| PC_32.0 | 5.21E-01 | 5.04E-01 | -0.107 | 1.48E-01 | 3.84E-01 |
| biogenic amine | 6.40E-02 | 1.64E-04 | -0.107 | 1.48E-01 | 3.84E-01 |
| TG_53.4 | 4.33E-01 | 2.00E-01 | -0.107 | 1.49E-01 | 3.84E-01 |
| PC_34.1 | 9.01E-01 | 5.04E-02 | -0.105 | 1.53E-01 | 3.91E-01 |
| PC_39.6 | 6.16E-01 | 8.37E-01 | -0.105 | 1.54E-01 | 3.91E-01 |
| TG_44.2 | 5.23E-02 | 2.04E-01 | -0.105 | 1.57E-01 | 3.91E-01 |
| PC_39.7 | 7.25E-01 | 4.89E-03 | -0.104 | 1.57E-01 | 3.91E-01 |
| DG_42.0 | 2.20E-04 | 2.36E-03 | -0.103 | 1.64E-01 | 4.04E-01 |
| PC_40.7 | 7.31E-01 | 5.51E-01 | -0.102 | 1.65E-01 | 4.04E-01 |
| PC_36.6 | 8.35E-01 | 2.78E-01 | -0.102 | 1.68E-01 | 4.06E-01 |
| AC_0.0 | 3.45E-01 | 8.74E-03 | -0.101 | 1.69E-01 | 4.06E-01 |
| PC_O_34.3 | 2.24E-01 | 5.80E-01 | -0.101 | 1.73E-01 | 4.12E-01 |
| Asn | 1.96E-05 | 1.14E-05 | -0.098 | 1.86E-01 | 4.38E-01 |
| Gly | 8.09E-01 | 3.66E-02 | -0.096 | 1.91E-01 | 4.47E-01 |
| SM_40.1 | 1.30E-11 | 1.88E-11 | -0.095 | 1.96E-01 | 4.48E-01 |
| SM_34.2 | 3.00E-01 | 1.44E-02 | -0.095 | 1.99E-01 | 4.48E-01 |
| Lys | 1.47E-02 | 8.04E-01 | -0.095 | 1.99E-01 | 4.48E-01 |
| DG_44.3 | 1.85E-06 | 2.37E-01 | -0.095 | 1.99E-01 | 4.48E-01 |
| PC_32.2 | 4.76E-02 | 6.08E-02 | -0.095 | 2.00E-01 | 4.48E-01 |
| PC_34.3 | 4.88E-01 | 7.57E-01 | -0.094 | 2.03E-01 | 4.48E-01 |
| AC_10.1 | 2.79E-04 | 1.44E-12 | -0.094 | 2.03E-01 | 4.48E-01 |
| PC_32.1 | 7.28E-04 | 7.45E-04 | -0.093 | 2.10E-01 | 4.52E-01 |
| PC_36.2 | 6.97E-01 | 9.01E-03 | -0.093 | 2.10E-01 | 4.52E-01 |
| CE_18.1 | 6.05E-01 | 3.82E-01 | -0.093 | 2.10E-01 | 4.52E-01 |
| SM_36.2 | 8.68E-01 | 6.20E-02 | -0.092 | 2.15E-01 | 4.59E-01 |
| PC_38.4 | 6.96E-01 | 2.21E-01 | -0.091 | 2.18E-01 | 4.62E-01 |
| LPC_16.0 | 2.80E-01 | 3.22E-01 | -0.090 | 2.23E-01 | 4.68E-01 |
| PC_O_38.3 | 1.27E-12 | 1.28E-13 | -0.089 | 2.27E-01 | 4.73E-01 |
| DG_42.2 | 5.39E-02 | 5.08E-01 | -0.088 | 2.33E-01 | 4.81E-01 |
| TG_56.8 | 8.70E-02 | 2.12E-01 | -0.086 | 2.43E-01 | 5.00E-01 |
| PC_39.3 | 5.42E-01 | 5.59E-02 | -0.085 | 2.48E-01 | 5.06E-01 |
| PC_40.3 | 1.83E-01 | 3.83E-03 | -0.084 | 2.54E-01 | 5.11E-01 |
| SM_30.1 | 1.65E-01 | 3.08E-02 | -0.084 | 2.56E-01 | 5.11E-01 |
| H1 | 1.49E-10 | 1.58E-04 | -0.084 | 2.57E-01 | 5.11E-01 |
| PC_38.2 | 2.41E-08 | 1.35E-08 | -0.083 | 2.63E-01 | 5.14E-01 |
| SM_32.1 | 3.15E-01 | 3.68E-01 | -0.083 | 2.63E-01 | 5.14E-01 |
| LPC_20.1 | 3.96E-02 | 8.00E-07 | -0.083 | 2.63E-01 | 5.14E-01 |
| PC_34.4 | 3.95E-02 | 3.17E-02 | -0.082 | 2.66E-01 | 5.14E-01 |
| PC_30.0 | 5.89E-01 | 4.39E-01 | -0.081 | 2.70E-01 | 5.20E-01 |
| LPC_O_18.2 | 3.60E-02 | 3.78E-08 | -0.081 | 2.73E-01 | 5.21E-01 |
| Serotonin | 1.47E-04 | 4.40E-11 | 0.080 | 2.76E-01 | 5.23E-01 |
| Val | 3.48E-01 | 4.01E-01 | -0.079 | 2.83E-01 | 5.27E-01 |
| Tyr | 3.86E-01 | 4.33E-01 | -0.079 | 2.86E-01 | 5.27E-01 |
| AC_10.3 | 4.24E-07 | 1.11E-05 | -0.079 | 2.86E-01 | 5.27E-01 |
| AC_8.0 | 6.93E-15 | 7.19E-14 | -0.079 | 2.86E-01 | 5.27E-01 |
| PC_O_36.4 | 2.58E-01 | 5.59E-03 | -0.079 | 2.88E-01 | 5.27E-01 |
| TG_44.1 | 2.19E-05 | 1.03E-04 | -0.078 | 2.90E-01 | 5.27E-01 |
| TG_53.5 | 2.73E-01 | 7.28E-01 | -0.078 | 2.94E-01 | 5.31E-01 |
| PC_36.4 | 3.52E-01 | 5.98E-01 | -0.077 | 3.00E-01 | 5.39E-01 |
| PC_37.6 | 6.44E-01 | 8.03E-01 | -0.076 | 3.06E-01 | 5.46E-01 |
| AC_8.1 | 3.32E-06 | 3.93E-07 | -0.075 | 3.09E-01 | 5.47E-01 |
| SM_40.2 | 1.87E-06 | 9.14E-11 | -0.074 | 3.16E-01 | 5.55E-01 |
| PC_O_32.1 | 3.10E-01 | 6.59E-02 | -0.074 | 3.18E-01 | 5.56E-01 |
| PC_36.3 | 8.05E-01 | 3.20E-02 | -0.073 | 3.23E-01 | 5.62E-01 |
| DG_O_36.4 | 1.02E-02 | 4.54E-01 | -0.072 | 3.32E-01 | 5.74E-01 |
| SM_34.1 | 6.88E-01 | 2.16E-01 | -0.071 | 3.36E-01 | 5.76E-01 |
| Met | 3.39E-03 | 9.21E-02 | -0.071 | 3.40E-01 | 5.78E-01 |
| PC_31.0 | 3.68E-01 | 3.23E-02 | 0.070 | 3.41E-01 | 5.78E-01 |
| PC_40.2 | 1.59E-08 | 4.14E-06 | -0.070 | 3.45E-01 | 5.81E-01 |
| Met_SO | 1.15E-16 | 2.08E-05 | 0.069 | 3.49E-01 | 5.83E-01 |
| PC_35.4 | 3.18E-01 | 4.73E-01 | -0.066 | 3.72E-01 | 6.16E-01 |
| PC_38.3 | 1.71E-08 | 2.60E-10 | -0.066 | 3.73E-01 | 6.16E-01 |
| PC_37.5 | 6.71E-02 | 3.31E-01 | -0.066 | 3.75E-01 | 6.16E-01 |
| PC_O_42.4 | 3.49E-05 | 3.06E-04 | -0.065 | 3.79E-01 | 6.19E-01 |
| AC_14.1 | 1.38E-08 | 2.33E-07 | 0.065 | 3.82E-01 | 6.20E-01 |
| PC_44.10 | 1.51E-11 | 9.26E-09 | 0.064 | 3.89E-01 | 6.23E-01 |
| TG_44.4 | 6.24E-10 | 1.69E-09 | -0.064 | 3.89E-01 | 6.23E-01 |
| His | 1.23E-15 | 4.47E-02 | -0.063 | 3.94E-01 | 6.26E-01 |
| LPC | 5.70E-01 | 9.69E-02 | -0.063 | 3.96E-01 | 6.26E-01 |
| xLeu | 5.72E-01 | 1.44E-01 | -0.063 | 3.98E-01 | 6.26E-01 |
| LPC_17.1 | 9.00E-01 | 5.40E-02 | 0.062 | 4.00E-01 | 6.26E-01 |
| PC_44.1 | 3.39E-15 | 7.38E-16 | 0.062 | 4.02E-01 | 6.26E-01 |
| Glu | 4.18E-01 | 1.26E-02 | -0.061 | 4.08E-01 | 6.31E-01 |
| PC_O_42.6 | 8.83E-07 | 3.83E-06 | -0.061 | 4.12E-01 | 6.34E-01 |
| AC_14.0_OH | 4.72E-03 | 4.43E-02 | -0.060 | 4.17E-01 | 6.38E-01 |
| PC_O_34.2 | 5.56E-01 | 5.15E-01 | -0.059 | 4.21E-01 | 6.41E-01 |
| TG_55.6 | 1.11E-08 | 1.16E-08 | -0.059 | 4.25E-01 | 6.44E-01 |
| PC_O_34.4 | 2.29E-01 | 7.59E-01 | -0.058 | 4.30E-01 | 6.44E-01 |
| AC_4.0 | 3.54E-12 | 3.16E-08 | -0.058 | 4.30E-01 | 6.44E-01 |
| PC_O_40.5 | 7.26E-01 | 9.91E-03 | -0.057 | 4.42E-01 | 6.58E-01 |
| SM_33.2 | 9.01E-02 | 1.63E-03 | -0.056 | 4.45E-01 | 6.59E-01 |
| PC_35.5 | 6.47E-03 | 8.55E-01 | -0.055 | 4.60E-01 | 6.78E-01 |
| Cer_41.1 | 8.06E-01 | 1.02E-01 | -0.054 | 4.66E-01 | 6.82E-01 |
| PC_O_40.4 | 3.13E-02 | 1.19E-02 | 0.054 | 4.68E-01 | 6.82E-01 |
| CE_16.0 | 3.63E-01 | 7.34E-01 | -0.052 | 4.79E-01 | 6.94E-01 |
| SM_37.1 | 7.70E-01 | 5.68E-01 | -0.051 | 4.89E-01 | 7.04E-01 |
| Taurine | 1.11E-03 | 1.67E-02 | -0.051 | 4.91E-01 | 7.04E-01 |
| SM_42.1 | 1.16E-02 | 9.14E-05 | -0.050 | 4.95E-01 | 7.04E-01 |
| PC_O_30.0 | 2.91E-01 | 5.78E-02 | -0.050 | 4.96E-01 | 7.04E-01 |
| AC_14.2 | 3.74E-10 | 1.54E-09 | 0.047 | 5.21E-01 | 7.33E-01 |
| LPC_20.4 | 9.54E-01 | 5.30E-01 | -0.047 | 5.22E-01 | 7.33E-01 |
| PC_37.4 | 6.94E-01 | 6.46E-01 | -0.047 | 5.27E-01 | 7.36E-01 |
| AC_18.1 | 6.92E-06 | 3.77E-04 | -0.046 | 5.31E-01 | 7.37E-01 |
| alpha_AAA | 3.68E-01 | 5.70E-01 | -0.045 | 5.41E-01 | 7.48E-01 |
| AC_5.0 | 4.31E-15 | 3.31E-06 | -0.045 | 5.46E-01 | 7.51E-01 |
| LPC_16.1 | 4.11E-07 | 3.03E-04 | -0.044 | 5.51E-01 | 7.51E-01 |
| TG_49.1 | 5.86E-03 | 2.28E-03 | -0.044 | 5.52E-01 | 7.51E-01 |
| PC_39.4 | 7.72E-02 | 6.39E-01 | -0.044 | 5.54E-01 | 7.51E-01 |
| PC_30.1 | 4.35E-07 | 4.11E-07 | -0.043 | 5.63E-01 | 7.59E-01 |
| SM_40.4 | 1.80E-04 | 2.53E-02 | -0.042 | 5.71E-01 | 7.66E-01 |
| PC_37.3 | 1.42E-01 | 4.46E-01 | -0.042 | 5.74E-01 | 7.66E-01 |
| PC_O_38.4 | 5.90E-01 | 2.56E-01 | -0.041 | 5.81E-01 | 7.72E-01 |
| Cer_43.1 | 4.80E-01 | 7.37E-01 | -0.040 | 5.92E-01 | 7.83E-01 |
| PC_36.1 | 1.69E-07 | 2.24E-07 | -0.039 | 6.01E-01 | 7.90E-01 |
| PC_33.0 | 3.91E-02 | 8.04E-01 | 0.038 | 6.03E-01 | 7.90E-01 |
| Ser | 6.56E-01 | 4.14E-01 | -0.038 | 6.08E-01 | 7.93E-01 |
| CE_20.4 | 6.64E-01 | 4.46E-01 | -0.036 | 6.27E-01 | 8.13E-01 |
| AC_10.0 | 1.99E-07 | 1.01E-11 | -0.035 | 6.35E-01 | 8.19E-01 |
| PC_35.2 | 5.53E-01 | 4.76E-01 | -0.034 | 6.45E-01 | 8.21E-01 |
| PC_O_35.3 | 4.01E-02 | 1.07E-01 | -0.034 | 6.48E-01 | 8.21E-01 |
| Thr | 5.94E-01 | 7.33E-01 | -0.034 | 6.49E-01 | 8.21E-01 |
| Cit | 2.99E-04 | 2.32E-05 | -0.034 | 6.49E-01 | 8.21E-01 |
| CE_16.1 | 2.59E-01 | 6.50E-01 | -0.033 | 6.52E-01 | 8.21E-01 |
| AC_18.2 | 1.46E-06 | 3.14E-06 | -0.033 | 6.54E-01 | 8.21E-01 |
| AC_6.0 | 3.21E-16 | 1.17E-13 | -0.032 | 6.65E-01 | 8.29E-01 |
| Asp | 7.52E-01 | 3.66E-01 | 0.032 | 6.67E-01 | 8.29E-01 |
| LPC_22.5 | 5.98E-07 | 8.10E-11 | -0.031 | 6.75E-01 | 8.31E-01 |
| PC_33.3 | 3.34E-02 | 4.27E-03 | 0.031 | 6.75E-01 | 8.31E-01 |
| PC_33.2 | 5.46E-01 | 7.59E-01 | -0.031 | 6.79E-01 | 8.33E-01 |
| PC_37.2 | 4.69E-01 | 5.43E-01 | -0.030 | 6.89E-01 | 8.39E-01 |
| AC_13.0 | 7.61E-17 | 7.20E-16 | 0.029 | 6.91E-01 | 8.39E-01 |
| PC_O_36.2 | 2.60E-01 | 1.80E-01 | -0.029 | 6.97E-01 | 8.44E-01 |
| PC_O_42.5 | 5.29E-07 | 5.79E-05 | -0.028 | 7.05E-01 | 8.50E-01 |
| AC_12.1 | 1.55E-08 | 1.60E-08 | 0.027 | 7.16E-01 | 8.55E-01 |
| SM_35.1 | 4.48E-01 | 2.03E-01 | -0.027 | 7.16E-01 | 8.55E-01 |
| PC_38.7 | 1.05E-01 | 1.39E-02 | -0.026 | 7.28E-01 | 8.65E-01 |
| PC_O_37.6 | 2.32E-01 | 2.15E-02 | -0.025 | 7.31E-01 | 8.65E-01 |
| Phe | 3.92E-02 | 6.19E-02 | -0.024 | 7.42E-01 | 8.74E-01 |
| Putrescine | 2.25E-03 | 3.78E-11 | -0.023 | 7.54E-01 | 8.81E-01 |
| PC_37.1 | 8.60E-01 | 2.30E-01 | -0.023 | 7.57E-01 | 8.81E-01 |
| PC_O_32.3 | 2.12E-02 | 8.51E-09 | -0.023 | 7.58E-01 | 8.81E-01 |
| SM_31.1 | 2.30E-02 | 1.06E-03 | -0.023 | 7.61E-01 | 8.81E-01 |
| Cer_40.1 | 6.54E-02 | 4.51E-01 | -0.021 | 7.81E-01 | 9.01E-01 |
| PC_32.4 | 5.97E-01 | 1.76E-02 | -0.020 | 7.88E-01 | 9.05E-01 |
| LPC_17.0 | 5.13E-01 | 1.27E-01 | 0.019 | 7.99E-01 | 9.06E-01 |
| SDMA | 3.39E-02 | 6.43E-13 | 0.019 | 8.01E-01 | 9.06E-01 |
| Arg | 3.82E-01 | 4.89E-01 | 0.018 | 8.07E-01 | 9.06E-01 |
| AC_10.2 | 1.51E-17 | 1.14E-10 | -0.018 | 8.09E-01 | 9.06E-01 |
| LPC_18.1 | 4.94E-02 | 5.39E-01 | 0.018 | 8.09E-01 | 9.06E-01 |
| PC_40.5 | 5.75E-04 | 1.21E-01 | -0.018 | 8.10E-01 | 9.06E-01 |
| c4_OH_Pro | 1.33E-13 | 1.75E-10 | 0.017 | 8.14E-01 | 9.06E-01 |
| Orn | 3.48E-01 | 5.59E-01 | -0.017 | 8.16E-01 | 9.06E-01 |
| aminoacids | 4.61E-03 | 6.20E-01 | -0.016 | 8.34E-01 | 9.22E-01 |
| PC_40.8 | 3.17E-02 | 9.22E-02 | 0.014 | 8.48E-01 | 9.34E-01 |
| CE_17.1 | 1.29E-02 | 4.52E-01 | 0.013 | 8.65E-01 | 9.45E-01 |
| DG_32.2 | 1.06E-01 | 3.46E-02 | -0.013 | 8.65E-01 | 9.45E-01 |
| Ac_Orn | 6.07E-01 | 9.37E-03 | 0.012 | 8.76E-01 | 9.50E-01 |
| PC_41.4 | 4.84E-10 | 2.06E-10 | 0.012 | 8.76E-01 | 9.50E-01 |
| AC_16.1 | 6.39E-12 | 5.54E-09 | -0.010 | 8.92E-01 | 9.50E-01 |
| PC_35.3 | 8.87E-01 | 1.20E-01 | 0.010 | 8.92E-01 | 9.50E-01 |
| SM_39.2 | 1.98E-09 | 7.30E-06 | -0.010 | 8.93E-01 | 9.50E-01 |
| SM_31.0 | 3.11E-06 | 4.66E-10 | -0.010 | 8.95E-01 | 9.50E-01 |
| PC_39.5 | 1.56E-02 | 1.39E-01 | 0.010 | 8.97E-01 | 9.50E-01 |
| TG_51.1 | 2.08E-14 | 4.40E-12 | -0.010 | 8.97E-01 | 9.50E-01 |
| PC_42.10 | 1.25E-02 | 2.37E-01 | 0.009 | 9.03E-01 | 9.52E-01 |
| Spermidine | 7.77E-15 | 1.10E-13 | 0.008 | 9.10E-01 | 9.56E-01 |
| AC_18.0 | 2.38E-19 | 2.04E-18 | 0.008 | 9.16E-01 | 9.56E-01 |
| PC_42.6 | 1.27E-06 | 8.32E-08 | 0.008 | 9.19E-01 | 9.56E-01 |
| AC_2.0 | 6.60E-01 | 2.48E-01 | -0.007 | 9.21E-01 | 9.56E-01 |
| LPC_18.2 | 7.52E-02 | 6.42E-01 | -0.007 | 9.24E-01 | 9.56E-01 |
| AC_16.0 | 5.47E-18 | 1.51E-16 | -0.007 | 9.30E-01 | 9.58E-01 |
| AC_12.0 | 2.56E-18 | 2.27E-16 | -0.006 | 9.40E-01 | 9.65E-01 |
| LPC_15.0 | 1.30E-01 | 5.50E-02 | 0.005 | 9.47E-01 | 9.69E-01 |
| AC_14.0 | 4.52E-20 | 4.12E-18 | 0.004 | 9.60E-01 | 9.78E-01 |
| ADMA | 5.33E-05 | 5.54E-13 | 0.003 | 9.64E-01 | 9.79E-01 |
| PC_35.1 | 2.65E-01 | 6.94E-01 | 0.003 | 9.70E-01 | 9.81E-01 |
| TG_54.2 | 8.24E-09 | 6.87E-09 | -0.002 | 9.77E-01 | 9.84E-01 |
| CE_19.2 | 8.63E-01 | 5.69E-01 | -0.002 | 9.83E-01 | 9.86E-01 |
| LPC_O_18.0 | 3.27E-03 | 9.46E-06 | -0.001 | 9.87E-01 | 9.87E-01 |

| Feature | Normality benign | Normality malignant | Rg effect size | P-value | FDR |
|---|---|---|---|---|---|
| glrlm_RunPercentage | 7.01E-37 | 1.54E-18 | -0.603 | 6.45E-107 | 3.94E-105 |
| glrlm_LongRunEmphasis | 1.90E-41 | 4.26E-21 | 0.602 | 1.18E-106 | 3.94E-105 |
| glrlm_RunLengthNonUniformityNormalizec | 4.77E-34 | 4.53E-17 | -0.602 | 1.46E-106 | 3.94E-105 |
| glrlm_ShortRunEmphasis | 1.64E-37 | 9.45E-20 | -0.602 | 1.47E-106 | 3.94E-105 |
| glrlm_RunVariance | 5.57E-43 | 2.43E-21 | 0.600 | 3.77E-106 | 8.06E-105 |
| glszm_LargeAreaEmphasis | 6.00E-92 | 5.77E-42 | 0.594 | 4.63E-104 | 8.26E-103 |
| gldm_LargeDependenceEmphasis | 5.66E-54 | 1.68E-25 | 0.588 | 4.86E-102 | 7.35E-101 |
| glszm_ZonePercentage | 1.89E-14 | 3.22E-08 | -0.588 | 5.50E-102 | 7.35E-101 |
| gldm_DependenceNonUniformityNormaliz | 3.02E-37 | 1.64E-10 | -0.581 | 1.66E-99 | 1.97E-98 |
| glszm_ZoneVariance | 5.76E-93 | 7.07E-42 | 0.579 | 6.18E-99 | 6.61E-98 |
| gldm_SmallDependenceEmphasis | 2.39E-08 | 2.18E-03 | -0.579 | 1.07E-98 | 1.05E-97 |
| gldm_DependenceVariance | 6.84E-60 | 3.60E-26 | 0.546 | 4.12E-88 | 3.68E-87 |
| glcm_Imc2 | 9.47E-82 | 4.47E-32 | -0.536 | 4.82E-85 | 3.96E-84 |
| ngtdm_Strength | 2.71E-70 | 2.32E-26 | -0.532 | 1.36E-83 | 1.04E-82 |
| glszm_SizeZoneNonUniformityNormalized | 1.14E-17 | 1.66E-04 | -0.522 | 1.05E-80 | 7.06E-80 |
| glszm_SmallAreaEmphasis | 1.17E-17 | 1.47E-08 | -0.522 | 1.06E-80 | 7.06E-80 |
| glcm_Imc1 | 6.47E-27 | 1.18E-05 | 0.475 | 3.73E-67 | 2.35E-66 |
| glcm_Id | 1.86E-39 | 4.42E-20 | 0.453 | 3.69E-61 | 2.19E-60 |
| glcm_DifferenceAverage | 3.73E-70 | 3.18E-13 | -0.453 | 4.04E-61 | 2.28E-60 |
| glcm_Idm | 3.56E-43 | 2.56E-22 | 0.443 | 1.06E-58 | 5.67E-58 |
| glcm_Contrast | 7.79E-79 | 2.27E-30 | -0.437 | 4.66E-57 | 2.38E-56 |
| glcm_MCC | 4.99E-61 | 6.93E-09 | -0.434 | 1.85E-56 | 9.01E-56 |
| glcm_InverseVariance | 1.20E-46 | 6.14E-22 | 0.424 | 8.10E-54 | 3.77E-53 |
| ngtdm_Contrast | 6.42E-89 | 7.69E-36 | -0.416 | 6.09E-52 | 2.72E-51 |
| gldm_GrayLevelNonUniformity | 4.05E-86 | 5.14E-34 | 0.382 | 5.75E-44 | 2.46E-43 |
| glrlm_GrayLevelNonUniformity | 2.63E-86 | 3.16E-34 | 0.374 | 2.13E-42 | 8.75E-42 |
| glcm_DifferenceVariance | 2.19E-79 | 3.50E-32 | -0.368 | 6.10E-41 | 2.42E-40 |
| glcm_ClusterShade | 4.10E-87 | 5.44E-15 | -0.360 | 2.92E-39 | 1.12E-38 |
| shape_MeshVolume | 3.84E-85 | 1.29E-32 | 0.317 | 6.02E-31 | 2.22E-30 |
| glszm_GrayLevelNonUniformity | 5.30E-88 | 1.04E-38 | 0.301 | 6.24E-28 | 2.22E-27 |
| ngtdm_Busyness | 2.21E-76 | 2.62E-26 | 0.299 | 1.26E-27 | 4.34E-27 |
| shape_Flatness | 1.80E-12 | 2.64E-09 | 0.295 | 5.55E-27 | 1.86E-26 |
| shape_SurfaceVolumeRatio | 7.25E-94 | 9.98E-31 | -0.292 | 2.15E-26 | 6.97E-26 |
| glcm_SumSquares | 2.59E-80 | 1.71E-15 | -0.290 | 4.61E-26 | 1.45E-25 |
| shape_VoxelVolume | 3.82E-85 | 2.78E-35 | 0.288 | 7.94E-26 | 2.43E-25 |
| glcm_JointEnergy | 4.58E-75 | 4.27E-37 | -0.283 | 6.63E-25 | 1.97E-24 |
| gldm_DependenceEntropy | 4.11E-11 | 8.74E-19 | 0.282 | 8.08E-25 | 2.34E-24 |
| glrlm_RunLengthNonUniformity | 1.02E-85 | 1.97E-36 | 0.281 | 1.22E-24 | 3.44E-24 |
| glcm_JointEntropy | 1.19E-09 | 2.68E-16 | 0.272 | 4.19E-23 | 1.15E-22 |
| gldm_SmallDependenceHighGrayLevelEmp | 4.76E-80 | 5.73E-16 | -0.262 | 1.42E-21 | 3.80E-21 |
| shape_LeastAxisLength | 2.04E-72 | 4.11E-15 | 0.260 | 2.57E-21 | 6.70E-21 |
| glcm_Idmn | 5.48E-18 | 1.71E-08 | 0.258 | 5.76E-21 | 1.47E-20 |
| glcm_Idn | 5.14E-12 | 1.69E-02 | 0.241 | 1.51E-18 | 3.75E-18 |
| firstorder_Energy | 3.74E-86 | 6.09E-39 | 0.237 | 6.16E-18 | 1.50E-17 |
| glszm_ZoneEntropy | 3.23E-15 | 4.67E-19 | 0.236 | 6.96E-18 | 1.66E-17 |
| firstorder_TotalEnergy | 3.41E-86 | 1.15E-38 | 0.231 | 3.55E-17 | 8.25E-17 |
| firstorder_Range | 5.71E-67 | 4.43E-13 | -0.228 | 9.19E-17 | 2.09E-16 |
| firstorder_Skewness | 1.08E-35 | 9.41E-03 | -0.227 | 1.26E-16 | 2.80E-16 |
| glcm_Correlation | 1.58E-25 | 4.01E-16 | 0.225 | 2.78E-16 | 6.07E-16 |
| glcm_MaximumProbability | 1.60E-69 | 3.16E-35 | -0.222 | 6.50E-16 | 1.39E-15 |
| firstorder_Maximum | 3.83E-67 | 1.58E-13 | -0.215 | 4.63E-15 | 9.72E-15 |
| glcm_ClusterProminence | 2.37E-89 | 1.65E-29 | -0.211 | 1.40E-14 | 2.88E-14 |
| ngtdm_Coarseness | 2.19E-78 | 2.30E-36 | -0.210 | 2.02E-14 | 4.07E-14 |
| shape_SurfaceArea | 4.83E-89 | 1.45E-41 | 0.208 | 3.45E-14 | 6.84E-14 |
| ngtdm_Complexity | 2.04E-85 | 2.11E-26 | -0.205 | 7.76E-14 | 1.51E-13 |
| glcm_ClusterTendency | 9.46E-82 | 1.61E-06 | -0.204 | 1.01E-13 | 1.94E-13 |
| glszm_LargeAreaLowGrayLevelEmphasis | 1.45E-71 | 1.96E-38 | 0.204 | 1.18E-13 | 2.21E-13 |
| firstorder_Variance | 1.15E-80 | 4.01E-09 | -0.202 | 1.80E-13 | 3.33E-13 |
| gldm_GrayLevelVariance | 1.17E-80 | 3.75E-09 | -0.202 | 2.06E-13 | 3.74E-13 |
| glszm_SmallAreaHighGrayLevelEmphasis | 1.20E-79 | 3.79E-18 | -0.201 | 2.45E-13 | 4.32E-13 |
| glrlm_GrayLevelVariance | 1.41E-80 | 5.14E-09 | -0.201 | 2.46E-13 | 4.32E-13 |
| glszm_GrayLevelVariance | 4.45E-80 | 3.74E-07 | -0.196 | 8.15E-13 | 1.41E-12 |
| glcm_SumEntropy | 3.04E-28 | 6.44E-23 | 0.196 | 9.41E-13 | 1.60E-12 |
| firstorder_MeanAbsoluteDeviation | 1.96E-70 | 3.24E-13 | -0.173 | 2.60E-10 | 4.34E-10 |
| gldm_DependenceNonUniformity | 1.83E-88 | 2.36E-40 | 0.168 | 1.02E-09 | 1.67E-09 |
| shape_Elongation | 8.57E-31 | 4.43E-09 | 0.165 | 1.98E-09 | 3.21E-09 |
| gldm_SmallDependenceLowGrayLevelEmpl | 3.08E-42 | 1.20E-21 | -0.159 | 6.88E-09 | 1.10E-08 |
| firstorder_Kurtosis | 1.31E-70 | 4.08E-28 | -0.153 | 2.28E-08 | 3.58E-08 |
| shape_Maximum2DDiameterSlice | 7.82E-84 | 1.34E-38 | 0.152 | 3.09E-08 | 4.79E-08 |
| glszm_SizeZoneNonUniformity | 1.79E-88 | 1.59E-40 | 0.151 | 3.58E-08 | 5.48E-08 |
| glszm_HighGrayLevelZoneEmphasis | 1.42E-79 | 1.49E-14 | -0.126 | 4.34E-06 | 6.55E-06 |
| gldm_LargeDependenceLowGrayLevelEmp | 8.64E-68 | 2.87E-37 | 0.122 | 8.42E-06 | 1.25E-05 |
| firstorder_RobustMeanAbsoluteDeviation | 2.72E-69 | 2.43E-11 | -0.117 | 1.90E-05 | 2.78E-05 |
| glszm_LargeAreaHighGrayLevelEmphasis | 1.27E-91 | 8.66E-42 | 0.113 | 3.85E-05 | 5.56E-05 |
| shape_Maximum2DDiameterColumn | 4.94E-83 | 1.56E-38 | 0.109 | 6.74E-05 | 9.61E-05 |
| glrlm_ShortRunHighGrayLevelEmphasis | 1.28E-79 | 1.69E-11 | -0.106 | 1.14E-04 | 1.61E-04 |
| firstorder_RootMeanSquared | 2.34E-57 | 5.53E-03 | -0.101 | 2.46E-04 | 3.42E-04 |
| glrlm_HighGrayLevelRunEmphasis | 1.45E-79 | 8.57E-12 | -0.100 | 2.53E-04 | 3.48E-04 |
| firstorder_InterquartileRange | 1.21E-67 | 1.93E-10 | -0.099 | 3.20E-04 | 4.34E-04 |
| gldm_HighGrayLevelEmphasis | 1.41E-79 | 9.22E-12 | -0.098 | 3.55E-04 | 4.75E-04 |
| gldm_LargeDependenceHighGrayLevelEmp | 4.11E-88 | 2.06E-29 | 0.097 | 3.85E-04 | 5.08E-04 |
| shape_Sphericity | 7.72E-18 | 3.79E-15 | 0.095 | 5.46E-04 | 7.12E-04 |
| firstorder_90Percentile | 5.62E-71 | 1.91E-13 | -0.090 | 1.11E-03 | 1.43E-03 |
| shape_MinorAxisLength | 5.42E-83 | 1.17E-38 | 0.089 | 1.16E-03 | 1.48E-03 |
| glcm_JointAverage | 2.47E-63 | 1.75E-07 | -0.083 | 2.40E-03 | 2.98E-03 |
| glcm_SumAverage | 2.47E-63 | 1.75E-07 | -0.083 | 2.40E-03 | 2.98E-03 |
| shape_Maximum2DDiameterRow | 5.97E-83 | 2.55E-39 | 0.081 | 3.13E-03 | 3.85E-03 |
| glrlm_LongRunHighGrayLevelEmphasis | 2.28E-79 | 9.78E-14 | -0.078 | 4.25E-03 | 5.17E-03 |
| glcm_Autocorrelation | 1.65E-80 | 5.58E-12 | -0.077 | 4.88E-03 | 5.87E-03 |
| glrlm_RunEntropy | 3.09E-26 | 7.26E-21 | 0.074 | 7.25E-03 | 8.62E-03 |
| shape_Maximum3DDiameter | 2.59E-83 | 1.23E-39 | 0.070 | 1.08E-02 | 1.27E-02 |
| glcm_DifferenceEntropy | 1.27E-38 | 2.25E-22 | 0.067 | 1.44E-02 | 1.68E-02 |
| glszm_SmallAreaLowGrayLevelEmphasis | 2.56E-41 | 2.09E-21 | -0.067 | 1.47E-02 | 1.69E-02 |
| glszm_GrayLevelNonUniformityNormalized | 7.08E-56 | 2.84E-30 | -0.064 | 1.98E-02 | 2.26E-02 |
| firstorder_Mean | 3.84E-60 | 3.55E-05 | -0.053 | 5.22E-02 | 5.88E-02 |
| firstorder_10Percentile | 1.72E-48 | 3.95E-27 | 0.048 | 8.37E-02 | 9.33E-02 |
| firstorder_Minimum | 2.45E-42 | 8.38E-14 | 0.042 | 1.27E-01 | 1.40E-01 |
| glrlm_ShortRunLowGrayLevelEmphasis | 2.88E-40 | 5.97E-22 | -0.042 | 1.30E-01 | 1.42E-01 |
| shape_MajorAxisLength | 1.77E-84 | 1.54E-39 | -0.041 | 1.36E-01 | 1.47E-01 |
| gldm_LowGrayLevelEmphasis | 1.09E-41 | 2.52E-23 | -0.039 | 1.52E-01 | 1.62E-01 |
| glrlm_LowGrayLevelRunEmphasis | 4.89E-41 | 6.20E-23 | -0.037 | 1.82E-01 | 1.93E-01 |
| glrlm_GrayLevelNonUniformityNormalized | 2.17E-56 | 3.34E-30 | -0.031 | 2.56E-01 | 2.68E-01 |
| firstorder_Uniformity | 8.92E-57 | 3.11E-30 | -0.026 | 3.39E-01 | 3.52E-01 |
| glrlm_LongRunLowGrayLevelEmphasis | 2.55E-44 | 2.29E-26 | -0.017 | 5.27E-01 | 5.42E-01 |
| firstorder_Median | 1.08E-54 | 5.28E-11 | 0.016 | 5.72E-01 | 5.83E-01 |
| firstorder_Entropy | 2.06E-29 | 1.65E-20 | 0.012 | 6.65E-01 | 6.71E-01 |
| glszm_LowGrayLevelZoneEmphasis | 2.43E-38 | 2.46E-20 | -0.004 | 8.82E-01 | 8.82E-01 |